\documentclass{article}

\PassOptionsToPackage{numbers, compress}{natbib}

\usepackage[preprint]{neurips_2026}

\usepackage[utf8]{inputenc} 
\usepackage[T1]{fontenc}    
\usepackage{hyperref}       
\usepackage{url}            
\usepackage{booktabs}       
\usepackage{amsfonts}       
\usepackage{nicefrac}       
\usepackage{microtype}      
\usepackage{xcolor}         
\usepackage{amsmath}        
\usepackage{graphicx}       
\usepackage{fontawesome5}

\usepackage{xcolor}         
\usepackage{amsmath}        
\usepackage{graphicx}       
\usepackage{fontawesome5}

\definecolor{customPink}{HTML}{D81B60}
\hypersetup{
    colorlinks=true,
    citecolor=teal,
    linkcolor=teal,
}

\title{\faGlobe~ ATLAS:\\Active Theory Learning for Automated Science}

 \author{
   {No\'{e}mi \'{E}ltet\H{o}} \\
   Google DeepMind\\
   London, UK \\
   \texttt{noemielteto@deepmind.com} \\
   \And
   Nathaniel D. Daw \\
   Princeton University; Google DeepMind \\
   Princeton, NJ, USA \\
   \texttt{ndaw@deepmind.com} \\
   \AND
   Kimberly L. Stachenfeld \\
   Google DeepMind; Colombia University \\
   New York, NY, USA \\
   \texttt{stachenfeld@deepmind.com} \\
   \And
   Kevin J. Miller \\
   Google DeepMind; University College London \\
   London, UK \\
   \texttt{kevinjmiller@deepmind.com} \\
}

\begin{document}

\maketitle

\begin{abstract}

Advancing scientific understanding through mechanistic modeling requires posing the right experimental questions to yield maximally informative data. To automate this pursuit within cognitive science, we introduce ATLAS (Active Theory Learning for Automated Science), an active learning framework for the data-driven discovery of interpretable behavioral models. ATLAS iterates between generating mechanistic hypotheses—instantiated as a diverse ensemble of sparse neural networks (Disentangled RNNs)—and designing experiments that optimally distinguish between them. We test this approach on the problem of recovering reinforcement learning agents from their behavior in bandit tasks. ATLAS designs varied sequences of qualitatively novel experiments with temporal structure tailored to underlying agent characteristics. The models trained on these experiments are evaluated against a comprehensive set of metrics for mechanistic modeling that capture behavioral, structural, and computational similarity. ATLAS achieves a $5\text{--}10\times$ improvement in sample efficiency across all metrics compared to random experimentation, and its performance is further validated against expert-designed experiments derived from literature. These \textit{in silico} results showcase ATLAS's potential to accelerate human-interpretable insights in cognitive science and other domains where scientific inquiry relies on discovering mechanistic models.


\end{abstract}

\section{Introduction}

\begin{figure}[!h]
  \centering
  \includegraphics[width=1.\textwidth]{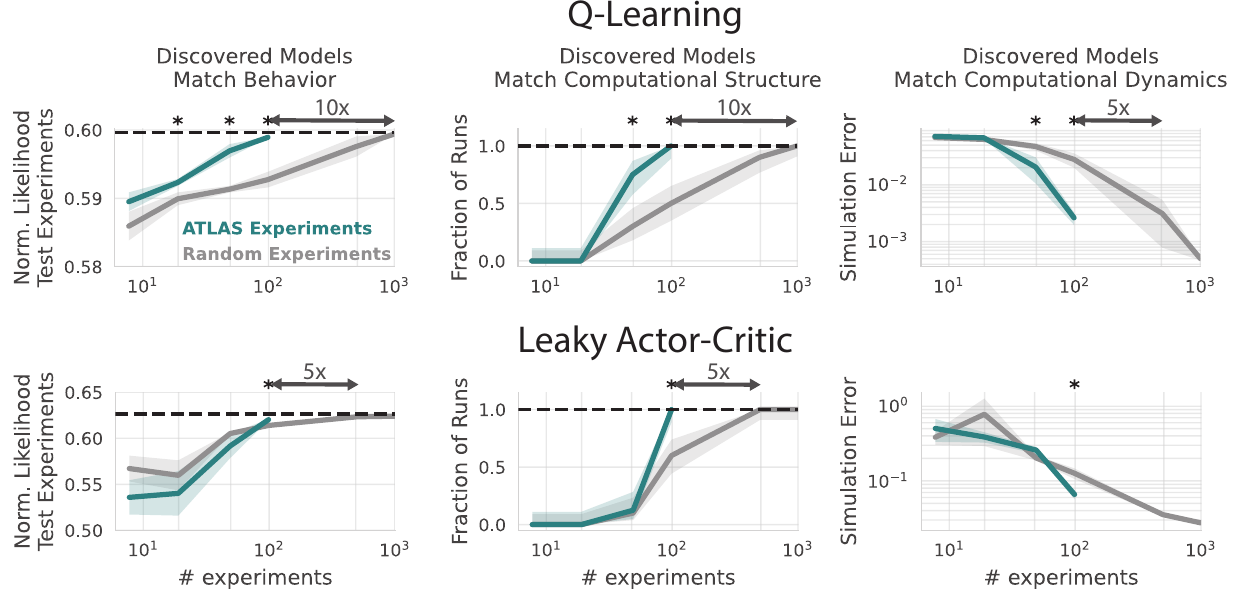}
  \caption{\textbf{ATLAS provides 5--10$\times$ sample efficiency over random experimentation}. Here, we use ATLAS to uncover the behavior and structure of two RL agents: Q-learning (Top) and Leaky Actor-Critic (Bottom). We compare 8 independent runs of ATLAS to 10 runs of random experimentation. \textbf{(Left)} Performance of the best-fitting models at predicting agent behavior in held-out experiments. Solid lines indicate the mean performance across independent runs, while the shaded regions represent $\pm 1$ standard error of the mean (SEM). Asterisks indicate significant advantage of ATLAS over random experiments ($p < 0.05$, of Welch's one-tailed t-test). The horizontal dashed line represents the performance ceiling. \textbf{(Middle)} The probability of recovering the true underlying computational graph. Solid lines denote the fraction of successful recoveries. Shaded regions represent the $68\%$  Wilson score interval. Asterisks indicate significance ($p < 0.05$, one-tailed Barnards's exact test). \textbf{(Right)} Averaged state prediction MSE for the ground truth agent simulated in the model and vice versa. Asterisks indicate significance ($p < 0.05$, Welch's one-tailed t-test).}
  \label{fig:eval}
\end{figure}

A central goal of cognitive science is to discover the algorithms that control the behavior of humans, animals, and artificial agents.
Research often proceeds via an iterative process of experimentation and computational modeling. Models serve as tentative hypotheses about the true algorithm, and guide the choice of next experiments. Data from experiments serves in turn to refine models \citep{box1976science}. This can be understood as an iterative closed-loop process alternating data-driven hypothesis discovery and hypothesis-driven data collection. Accelerating this process using machine learning remains a major open direction \citep{agrawal2020scaling, wang2023scientific, musslick2025automating, jagadish2026can}.

Recent techniques for data-driven model discovery make it possible to automatically identify computational models that both match data well and are interpretable to human researchers as plausible algorithmic hypotheses \citep{miller2023disrnn, brenner2024almost, ji2025discovering, rmus2025generating, castro2025discovering, lafollette2025data, eckstein2026hybrid, kasenberg2026ai}. These approaches surface a small set of models that are both simple and consistent with a dataset, from a vast space of possibilities. This generality comes with a cost: these methods typically require large datasets in order to identify meaningful models. This makes them difficult to apply in situations, typical in cognitive science, where experiments are costly and time-consuming.

Active Learning \citep{settles2009active} and Optimal Experiment Design \citep{huan2024optimal} are subfields of statistics and machine learning that are concerned with developing automated tools for designing experiments for efficient data collection. These tools iterate between identifying the most informative data points to sample and updating a model based on the new data.
While broadly applied, existing approaches typically either optimize black-box predictive models \citep{settles2009active, huan2024optimal}, which lack interpretability, or discriminate among pre-specified mechanistic models \citep{myung2009optimal, cavagnaro2010adaptive}, which carry a risk of misspecification and limit discovery.

To bridge this gap, we introduce ATLAS (Active Theory Learning for Automated Science). ATLAS implements an Active Learning loop using an ensemble of disentangled recurrent neural networks (DisRNNs) ~\citep{miller2023disrnn}, which have been used to discover computational models from behavior in a variety of contexts \citep{liu2024discovering, jain2025simultaneous, hoxha2025uncovering, zhu2025disentangling}. DisRNNs use information bottlenecks to encourage solutions that consist of a small number of sparsely interacting latent (``cognitive'') variables. The latent variable interactions can be interpreted as a hypothetical computational graph for the underlying mechanism. 
The strength of this sparsity constraint can be varied parametrically, and we use this to maintain ensemble diversity. We use an evolutionary algorithm to identify experimental parameters that optimize for disagreement among the members of the ensemble. 

We apply this approach to the problem of discovering mechanistic models of simple reinforcement agents based on their behavior, using as few behavioral experiments as possible (Figure \ref{fig:eval}). We find that ATLAS-designed experiments are able to identify strong models using $5\text{--}10\times$ fewer experiments than random experiments. These models are not only strong predictive models, accurately describing the behavior of the RL agents on out-of-distribution test experiments (Figure \ref{fig:eval},  left), but also discover the correct computational structure (Figure \ref{fig:eval}, middle) and dynamics (Figure \ref{fig:eval}, right). We also find that ATLAS matches or exceeds the performance of fixed, expert-designed baselines by tailoring its experiments to the specific agent being studied (Figure \ref{fig:handcrafted_vs_ATLAS}).

In addition to performing well, ATLAS-designed experiments often have interpretable structure: they exhibit a combination of regular and irregular structure on different timescales that make them effective at uncovering behavior, but quite unlike what researchers tend to design.

Our specific contributions are:
\begin{itemize}
    \item We demonstrate that combining methods from disagreement-based active learning with methods for automatic discovery of interpretable mechanistic models results in a practical framework for closed-loop scientific discovery. We refer to this framework, which links data-driven hypothesis generation with hypothesis-driven experiment selection, as Active Theory Learning for Automated Science (ATLAS).
    \item We demonstrate that Disentangled RNNs \citep{miller2018predictive} perform well in ATLAS, surfacing sufficiently diverse hypotheses to enable strong experimental design.
    \item We develop an optimizer over experiment designs that consist of binary matrices. We demonstrate that this results in experiments that drive strong disagreement between individual handcrafted models, and that it performs well in the context of ATLAS.
    \item We apply ATLAS to the problem of discovering reinforcement learning algorithms (Q-learning and Actor-Critic) from their overt behavior. We demonstrate that ATLAS discovers strong models of these algorithms using an order of magnitude fewer experiments than random experimentation, and that it outperforms even expert-designed experiments.
    
\end{itemize}

  \section{Related Work}

  \textbf{Active learning} studies how a learner should select the most informative data point to query \citep{settles2009active}. A foundational approach in this space is \emph{Query by Committee} (QBC) \citep{seung1992query,freund1992information}, in which a committee of models is maintained, each consistent with the observed data, and the learner queries the instance on which the committee members most disagree.
  QBC provides a natural framework for model-driven experimentation: the committee's disagreement identifies regions where the current data are insufficient to distinguish between competing hypotheses. Our approach shares this spirit, but replaces the committee with an ensemble of neural networks whose structure encourages qualitatively distinct models.
  
  Several measures of committee disagreement have been proposed. Vote entropy~\citep{dagan1995committee} and KL-divergence-based measures~\citep{mccallum1998employing} quantify the diversity of predictions across
  committee members. More recently, Bayesian Active Learning by Disagreement (BALD)~\citep{houlsby2011bayesian} formalized disagreement as the mutual information between model predictions and model parameters, providing a
  principled information-theoretic objective.
  
  \textbf{Optimal Experiment Design} applies Active Learning to science where the best experiment is the one that is most informative about a scientific theory, written as a computational model \citep{ryan2016review, rainforth2024modern}. Indeed, the expected information gain about the parameters of a scientific model \citep{lindley1956measure, mackay1992information, chaloner1995bayesian, foster2021deep} is theoretically unified with BALD \citep{kirsch2022unifying}.

  Optimal experiment design has been applied to numerous prediction and design problems, including materials discovery~\citep{lookman2019active}, drug design~\citep{warmuth2003active}, physics simulations ~\citep{baydin2021toward}, and cognitive science \citep{godara2026adversarial}. Notably, these applications involve fitting blackbox predictive models rather than identifying interpretable mechanistic models.
  
  In cognitive science and psychology, automated experiment design has been developed for parameter estimation of a fixed model ~\citep{myung2013tutorial, ouyang2016practical}, as well as for discriminating among model classes~\citep{myung2009optimal, cavagnaro2010adaptive}. However, as \citet{rainforth2024modern} note, ``experiment design is only ever as good as the underlying model.'' If the assumed model is fundamentally misspecified, it can drive a vicious cycle of biased data collection in irrelevant corners of the design space; escaping this trap remains an open challenge. Indeed, complex ground-truth models and/or restrictive model classes sometimes result in situations where random experiments outperform those selected by active learning \citep{dubova2026against, musslick2023evaluation}. This motivates approaches that reason over the model structure itself. Our work addresses this by leveraging structurally constrained neural networks, whose internal diversity naturally encodes qualitatively distinct mechanistic hypotheses.
  
\paragraph{Interpretable recurrent neural network models.}
Our approach builds on the Disentangled RNN (DisRNN; \cite{miller2023disrnn}), which learns recurrent dynamics with sparse, disentangled latent states. By penalizing superfluous internal representations, DisRNNs converge on tunably minimal models, ensuring that ensemble disagreement reflects meaningful structural differences rather than superficial parameter variation. This distinguishes our approach from standard deep ensemble methods \citep{lakshminarayanan2017simple}, where disagreement may arise from random initialization without corresponding to interpretably different hypotheses.

\section{The ATLAS pipeline}
\label{sec:atlas_pipeline}

\begin{figure}[h!]
  \centering
  \includegraphics[width=0.51\textwidth]{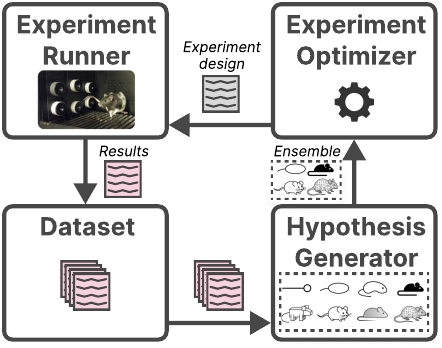}
  \caption{\textbf{Schematic of ATLAS.} An ATLAS cycle begins with either an initial, very small, dataset, from which it generates an ensemble of mechanistic models (Hypothesis Generator) or with a fixed set of models provided as input. Next it maximizes disagreement among the models to optimize the information gain of the experiment design (Experiment Optimizer). Finally, it runs that experiment (Experiment Runner) to expand the continually growing Dataset.}
  \label{fig:schematic}
\end{figure}

ATLAS consists of iterative cycles of model discovery and experimental design (Figure~\ref{fig:schematic}). We consider problems that involve discovering the computational structure of an unknown reinforcement learning agent, based on observations of its behavior on context-free bandit problems with binary rewards. In each experiment the agent chooses repeatedly one of $A$ available actions for up to $T$ trials and observes a binary reward. Different agents will adapt their future choices in response to the rewards in ways that reflect different computational structures and assumptions about the environment. These problems are close \textit{in silico} analogs of the problem of discovering the algorithms of human and animal reward learning from behavior, a widely-studied open problem in cognitive science \citep{collins2019reinforcement, eckstein2021what}.

\subsection{Experiment Optimizer}
\label{subsec:experiment_optimizer}

Generally, an experiment design $\mathbf{D}$ can be expressed as a tensor of time-varying inputs to the system of study. In our setting, we constrain $\mathbf{D}$ to be a binary $T{\times}A$ matrix $\mathbf{D}$, where each entry $\mathbf{D}_{t}(a)$ defines the reward received by an agent taking action $a$ at timestep $t$.

Each Hypothesis Generation step yields an ensemble $\mathcal{E}$ of $K$ mechanistic models. These can be thought of as competing hypotheses, plausible based on the data so far, for the generative reinforcement learning agent that ATLAS is trying to discover.
Each model $\mathcal{E}^i$ consists of an initial state vector $s_0$ and an update rule $s_{t+1}, \pi_{t+1} = F(s_t, a_t, r_t)$, which takes in the current state $s_t$, action $a_t$, and reward $r_t$ and outputs the updated state $s_{t+1}$ and the next action probabilities $\pi_{t+1}(a)$. 
Given a candidate experiment $\mathbf{D}$ and a model $\mathcal{E}^i$, simulated choice behavior can be sampled from the model. At each timestep $t$, upon the agent's choice $a_{t}$, the resulting reward is obtained from the experiment as $r_{t}=\mathbf{D}_{t}(a_{t})$.
This yields a simulated trajectory $\tau^i = (a^i_1, r^i_1, \ldots, a^i_T, r^i_T)$. To reduce variance, we generate a batch $\mathcal{B}$ of $B$ independent trajectories for each model, with $\tau_b^i$ indicating the $b^\text{th}$ trajectory sampled under model $\mathcal{E}_i$.
%

To compute model disagreement, we first compute the log-likelihood of the simulated trajectories under each model:

  \begin{equation}
    \ell_{ij}
    = \hat{\mathbb{E}}_{\mathcal{B}} \left[ \log p(\tau^i \mid \mathcal{E}^j)  \right]
    = \frac{1}{B}\sum_{b=1}^{B}\sum_{t=1}^{T}\log\pi^j_t(a^i_t|a_{1:t-1},r^i_{1:t-1})
  \end{equation} 
where $\ell_{ij}$ gives the average log-likelihood under the model $\mathcal{E}^j$ of a batch of $B$ trajectories $\tau^i$ generated by model $\mathcal{E}^i$.

  %

The expected information gain (EIG) from an experiment design $\mathbf{D}$ is approximated by the ensemble disagreement -- that is, by how much more likely each model considers its own trajectories compared to the trajectories of the other models:
  \begin{equation}\label{eq:eig}
    \mathrm{EIG}(d)
    = \frac{1}{K(K{-}1)}
      \sum_{i=1}^{K}\sum_{\substack{j=1\\j\neq i}}^{K}
      \bigl(\,
        \underbrace{\ell_{ii}}_{\text{self-agreement}}
        \;-\;
        \underbrace{\ell_{ij}}_{\text{cross-agreement}}
      \,\bigr).
  \end{equation}
$\mathrm{EIG}(d)$ is high when individual ensemble members predict distinct behavioral trajectories with high confidence, leading to strong disagreement among the models. Analogous to the BALD \citep{houlsby2011bayesian} objective, this score quantifies the expected information gain about the true identity of the generative model given the experimental outcome. Identifying a high-quality experiment for a given ensemble therefore requires finding an input $\mathbf{D}$ that maximizes this disagreement, thereby making the underlying models highly distinguishable.

We optimize Eq.~\ref{eq:eig} using an iterative hill climbing strategy. We initialize the binary experiment design matrix $\mathbf{D}$ randomly. We then consider elements one-at-a-time in a random order without replacement, flipping the associated bit and accepting the change if it increases the EIG and rejecting it otherwise. Once all elements have been considered we repeat this process, until a full pass through all entries yields no improvement, indicating that hill climbing has converged. To mitigate the risk of converging to a local optimum, we repeat this process 128 times using different randomly initialized experiments. We evaluate each converged experiment on a new set of rollouts $B'$ and select the experiment with the highest score. 

We first evaluate the Experiment Optimizer in isolation from the rest of ATLAS to demonstrate its sensitivity to differences within an ensemble. For this, we set up a simple example where the ensemble members are identical Q-learning agents that differ only in a single parameter: the learning rate. Specifically, we fixed a reference agent $Q_{\text{Medium}}$ ($\alpha=0.2$) and paired it with one of four comparison agents spanning slower ($Q_{\text{Very Slow}}$, $\alpha=0.05$; $Q_{\text{Slow}}$, $\alpha=0.1$) and faster ($Q_{\text{Fast}}$, $\alpha=0.4$; $Q_{\text{Very Fast}}$, $\alpha=0.8$) learning rates.

\begin{figure}[h!]
  \centering
  \includegraphics[width=1.\textwidth]{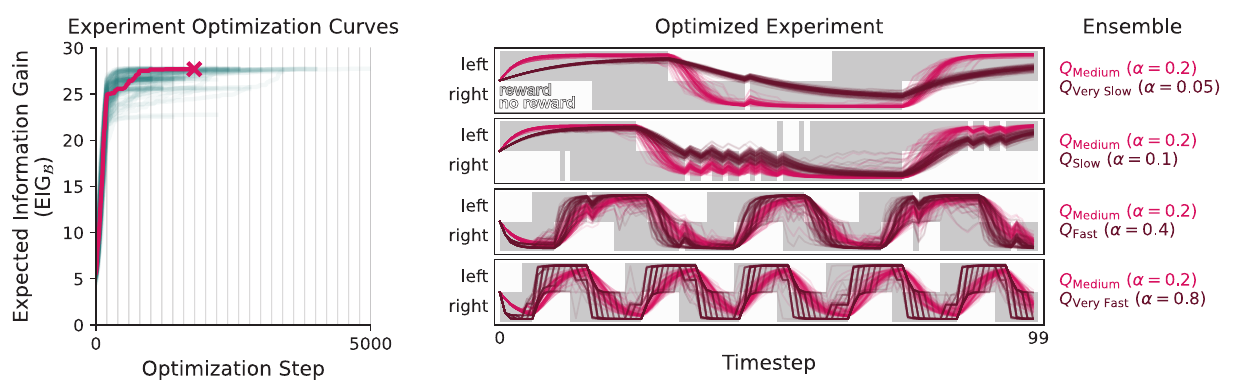}
  \caption{\textbf{Optimized experiments for Q-Learners with different learning rates}. We optimized binary reward matrices to distinguish a reference agent $Q_{\text{Medium}}$ with learning rate $\alpha=0.2$ from four comparison agents spanning slower and faster learning rates $\alpha=0.05$ to $\alpha=0.8$. \textbf{(Left)} For the $Q_{\text{Medium}}$ and $Q_{\text{Very Slow}}$ pair, the expected information gain (EIG) converges in $<5{,}000$ evolution timesteps. Thin lines in teal correspond to 128 seeds optimized on a batch of simulated agent trajectories $B=10 \times 1024$, and the thick line in pink is the experiment with the highest EIG as evaluated on a separate batch $B'=10 \times 1024$. Vertical lines represent convergence checks. \textbf{(Right)} The four panels display the optimized experiments for four pairs of Q-learning agents. Experiments consist of a binary reward matrix where gray indicates that a reward of 1 will be given depending on the action, and white indicates a reward of 0. Superimposed on each experiment is a small batch of 100 trajectories generated from each agent, with the action probabilities from the fixed reference agent ${Q_{\text{Medium}}}$ in pink, and the comparison agent ($Q_{\text{Very Slow}}$, $Q_{\text{Slow}}$, $Q_{\text{Fast}}$, or $Q_{\text{Very Fast}}$) in burgundy.}

  \label{fig:experiment_optimizer_timescales}
\end{figure}

Each of the four optimized experiments shows remarkable temporal structure (Figure \ref{fig:experiment_optimizer_timescales}). For example, the experiment optimized to distinguish $Q_{\text{Medium}}$ from $Q_{\text{Very Slow}}$ features a partially overlapping block structure: initially, only one choice is rewarding, followed by both, then the alternative choice, both again, and finally returning to the first. Notably, this pattern is interrupted by an anomalous reward omission. This omission must necessarily improve EIG; otherwise, the optimization algorithm would have flipped it prior to convergence. Indeed, this specific trial was part of the optimized design in all 10 independent runs (Appendix \ref{app:experiment_optimizer_seed_robustness}).

Examining the timeseries of choice probabilities from $Q_{\text{Medium}}$ and $Q_{\text{Very Slow}}$, we see that these anomalous trials cause a distinct fast or slow change in policy. We also see that the timing of the block changes is not arbitrary -- the blocks change when the policies of the two agents become similar. The other three experiments—optimized to distinguish $Q_{\text{Slow}}$, $Q_{\text{Fast}}$, and $Q_{\text{Very Fast}}$ from the reference—are temporally scaled versions of the first experiment. The frequency of alternation in these block designs maps faithfully onto the difference in learning timescales between the agents, confirming that the Experiment Optimizer successfully exploits meaningful ensemble diversity to maximize information gain, and the designs can be interpreted post-hoc.

\subsection{Experiment Runner and Dataset}
The Dataset is initially seeded with two random experiments. In each subsequent ATLAS cycle, the newly optimized experiment is executed, yielding a single behavioral trajectory from the ground-truth agent that is appended to the Dataset.

\begin{figure}[t!]
  \centering
  \includegraphics[width=1.\textwidth]{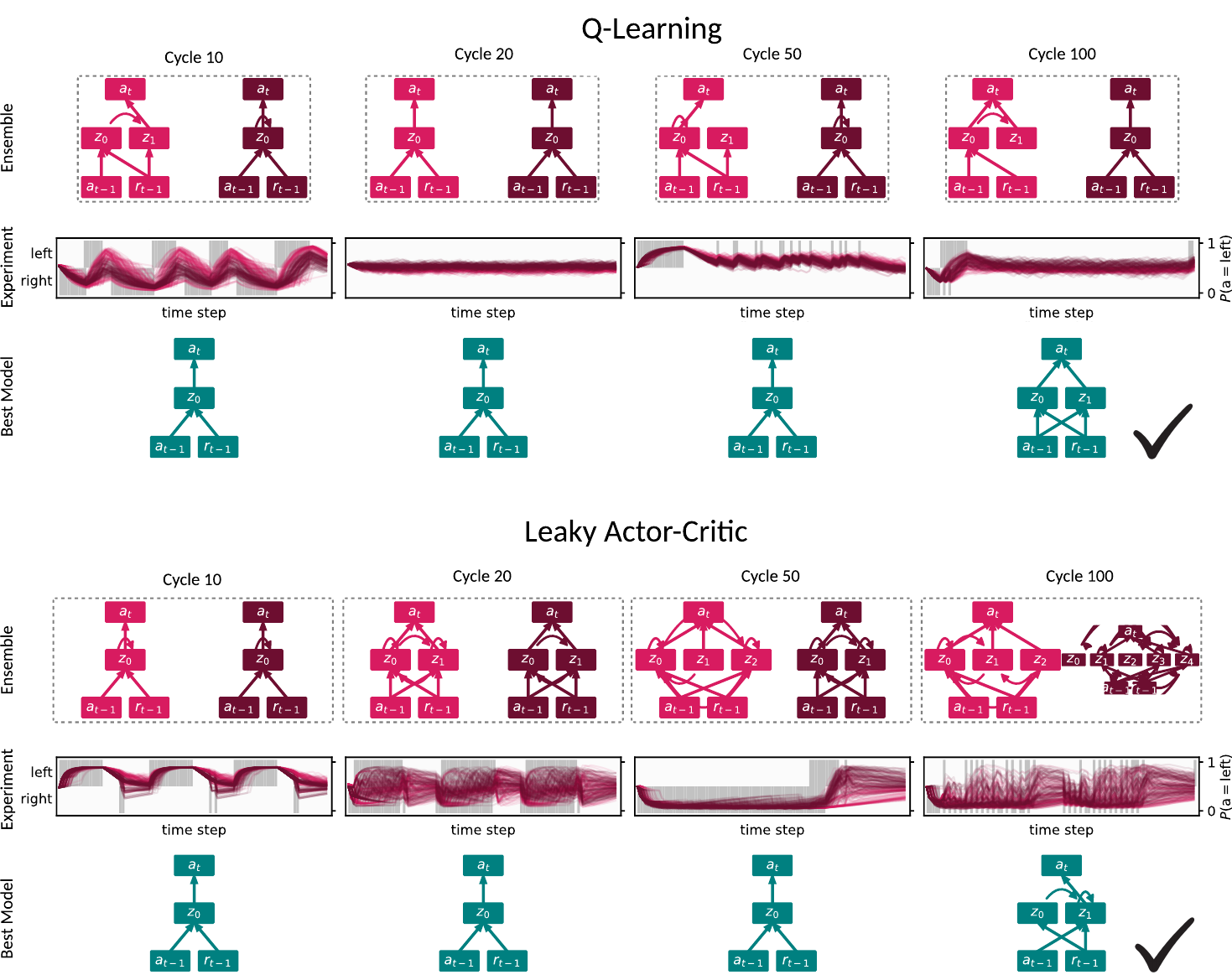}
    \caption{\textbf{Structured models drive structured experiments in ATLAS.} Example cycles from an arbitrarily chosen run (seed 2) are shown for Q-learning (top) and Leaky Actor-Critic (bottom). For each agent: (\textbf{First Row}) Computational graphs of the two ensemble members. (\textbf{Second Row}) Optimized experiments and $100$ simulated trajectories of the two ensemble members. There is rich temporal structure within each experiment, as well as a variety of structures over the sequence of experiments. (\textbf{Third Row}) Graph representations of the best-fit models, reflecting the shift from over-sparse approximations to a complex model that is isomorphic to the ground truth graph by cycle 100. Tick marks indicate isomorphism with the ground truth.}
  \label{fig:example_run}
\end{figure}

\subsection{Hypothesis Generator}
\label{subsec:hypothesis_generation}

Once the new experiment and its results have been added to the dataset, the next step in the ATLAS cycle is to create a new ensemble of hypotheses, instantiated as sparse neural networks. This involves two stages: First we generate a large and diverse set of candidate hypotheses, then we choose a subset of these to be used as the next ensemble of interest.

Specifically, the Hypothesis Generator trains a new set of Disentangled RNNs (DisRNNs, \citep{miller2023disrnn}) from scratch on the updated dataset.
The set includes both multiple random seeds as well as a range of values for the DisRNN penalty parameter, which controls the tradeoff between quality-of-fit and model complexity (Appendix \ref{app:disrnn_sweep}). For each set of hyperparameters, we train three networks: two networks on the two halves of the dataset (even and odd-numbered experiments), and one on the entire dataset.
We evaluate quality-of-fit for a set of hyperparameters using cross-validation between the even and odd splits, and take the model trained on the entire dataset as the hypothesis discovered with those hyperparameters.

This results in a large set of models with different tradeoffs between quality-of-fit and simplicity. Some of these, with higher values of the complexity penalty, will represent hypotheses that are too structurally simple to capture the existing experimental data, and will have poor quality of fit due to underfitting. Others, with lower values of the complexity penalty, will be too complex, and have poor quality of fit due to overfitting (Figure \ref{fig:ensemble_selection}). Models with good cross-validated quality of fit may be plausible mechanistic models given the current dataset. To select an ensemble of these, we sample models probabilistically using an \textit{offset softmax} distribution over candidate cross-validation likelihoods (Appendix \ref{app:ensemble_selection}). 
This has the effect that ATLAS is more exploratory on early cycles when model quality is poor, and more strongly prefers performant hypotheses in later cycles when quality is higher (Appendix \ref{app:ensemble_selection}). ATLAS is robust to other reasonable choices of ensemble selection, as uniform selection and top-K selection also yield competitive results (Appendix \ref{app:robustness_ensemble_selection}).

\section{Results}

\subsection{Experimental Setup}

We tested ATLAS separately with two ground-truth agents: a Q-learning agent and a Leaky Actor-Critic Agent \citep{sutton2018reinforcement} (see Appendix \ref{app:agents} for details). We selected these agents because they represent two different computational structures that both have relevance in cognitive science, and have previously been used as a test case for DisRNN \citep{miller2023disrnn}. To evaluate consistency, we performed eight independent ATLAS runs for each dataset, each for a total of 100 cycles. We compare ATLAS to two open-loop strategies for generating experiments: random rewards drawn i.i.d. or from Gaussian random walks.

Random experimentation is robust and widely applied when well-informed models are unavailable, from random mutagenesis in molecular biology \citep{nirantar2021directed}, to random screening in drug discovery \citep{vincent2022phenotypic}, to random neural perturbations in neuroscience \citep{chichilnisky2001, widloski2018inferring}. Within cognitive science, Musslick et al. \citep{musslick2023evaluation} clearly demonstrated cases where unbiased experimentation is superior to active learning, which tends to sample narrowly when guided by misspecified models. Therefore, our first baseline consists of a set of i.i.d. experiments where rewards on each trial are sampled independently with a probability of $0.5$.

Rather than relying on random designs, neuroscience and psychology labs typically study reward-guided learning and decision-making using fully handcrafted tasks or expert-defined procedural generators. A common example is to sample rewards from independent Gaussian random walks \citep{daw2006cortical, walton2010separable, miller2018predictive, eckstein2026hybrid}, which allows for substantial autocorrelation in reward probabilities and encourages the subject to develop strong preferences for one alternative or the other at various times in the experiment. We adopt this generator as our second baseline, matching the particular walks used in \cite{miller2018predictive}, an early dataset collected with the goal of supporting data-driven model discovery.

\subsection{Evaluation}
ATLAS is a tool for choosing which experiments to run, and its output is the resulting dataset. A dataset is successful if it is sufficient to allow discovering strong models of the ground-truth agent. We evaluate this by fitting a sweep of DisRNNs in a configuration that is optimized for discovery of a single strong model rather than for efficiently identifying a set of reasonable models (see Appendix \ref{app:disrnn_sweep} for details), then by evaluating the discovered model.

\begin{figure}[h!]
  \centering
  \includegraphics[width=1.\textwidth]{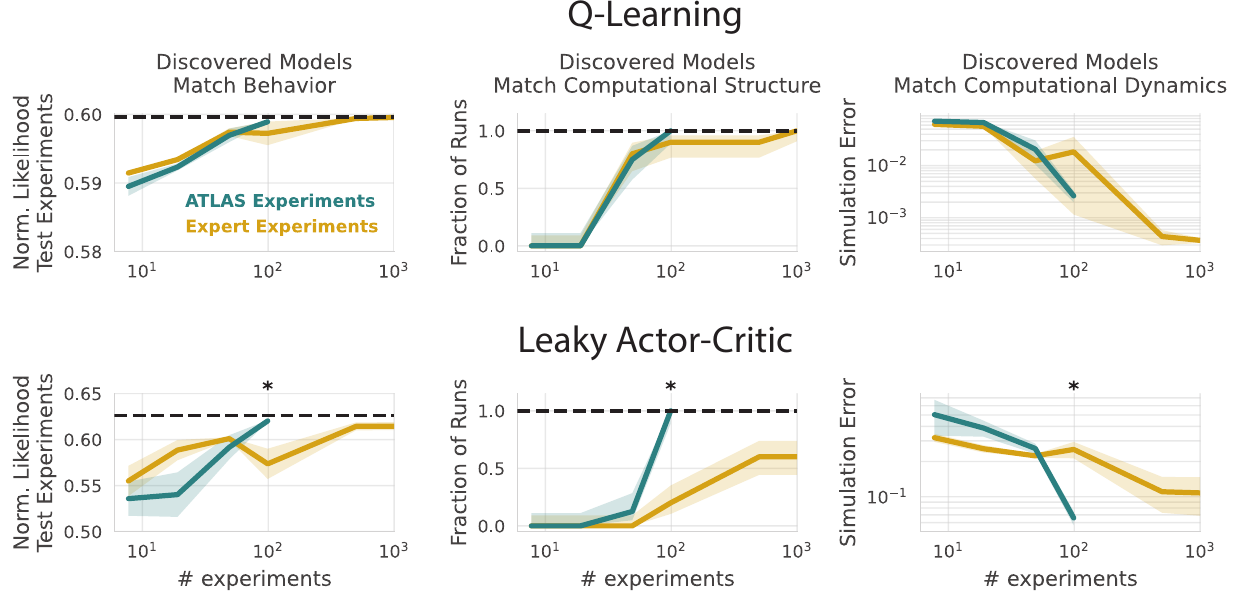}
  \caption{\textbf{ATLAS is competitive with expert-designed experiments.} We compare 8 independent runs for ATLAS, 10 runs for expert-designed on Q-learning, and 10 runs for expert-designed on leaky actor-critic. \textbf{(Left)} Performance of the best-fitting models at predicting agent behavior in held-out experiments. Solid lines indicate the mean performance across independent runs, while the shaded regions represent $\pm 1$ standard error of the mean (SEM) Asterisks indicate a significant advantage of ATLAS over expert-designed experiments ($p < 0.05$, Welch's one-tailed t-test). The horizontal dashed line represents the performance ceiling. \textbf{(Middle)} The probability of recovering the true underlying computational graph. Solid lines denote the fraction of successful recoveries. Shaded regions represent the $68\%$  Wilson score interval. Asterisks indicate significance ($p < 0.05$, one-tailed Barnards's exact test). \textbf{(Right)} Averaged state prediction error for the ground truth agent simulated in the model and vice versa. Asterisks indicate significance ($p < 0.05$, Welch's one-tailed t-test).}
  \label{fig:handcrafted_vs_ATLAS}
\end{figure}

We consider three criteria of strong mechanistic models. The first is \textit{behavioral similarity}: its behavior should be quantitatively like that of the system. We measure this using a test dataset of behavior from the ground-truth agent on a battery of handcrafted experiments designed to expose key patterns (see Appendix \ref{app:heldout_experiments} for details). We quantify the likelihood of this test dataset under the model. For strong models, this likelihood will approach that of the ground-truth agent itself. We find that ATLAS achieves high behavioral similarity with remarkable efficiency: after only 100 ATLAS-designed experiments, its performance nearly matches that of random search after 1,000 steps. Furthermore, when comparing equal milestones, ATLAS yields significantly higher behavioral similarity than random search after just 20 experiments for Q-learning, and after 100 experiments for Actor-Critic (Figure \ref{fig:eval}, left).

The second criterion is \textit{structural similarity}: whether there exists a 1:1 mapping between the components of the model and the components of the system. We represent the structure of each ground truth agent as a directed graph in which nodes are input variables (previous choice, previous reward), latent variables (for Q-learning: action values; for Leaky Actor-Critic: state value and policy parameters) or output variables (probability of next choice), and directed edges connect variables that causally affect one another (Figures \ref{fig:qlearning_graph} and \ref{fig:actorcritic_graph}). Similarly, a computational graph can be extracted from a fit DisRNN by examining the information bottlenecks, which encourage causal sparsity \citep{miller2023disrnn}. The nodes of this graph are similarly input variables, internal variables (activities of individual network units) and output variables, connected by directed edges when those variables are connected in the network by open information bottlenecks. After training the DisRNN parameters, including the fitted noise $\sigma$ for each bottleneck, we threshold the bottlenecks to be either ``open'' (information passes) if $\sigma<.5$ or ``closed'' (no information passes) otherwise. This thresholded DisRNN is structurally similar to the ground truth agent when their computational graphs are isomorphic. We find that ATLAS succeeds in recovering the correct computational graph for 8/8 seeds after 100 experiments for both the Q-learning and Leaky Actor-Critic agents. For the Q-learning agent, both the random and the expert-designed baselines required 1,000 experiments to reach this performance (Figure \ref{fig:eval}, middle; Figure \ref{fig:handcrafted_vs_ATLAS}, middle). For the Actor-Critic agent, the random baseline achieved this in 1,000 experiments, while the expert-designed baseline never achieved it.

The third and final criterion is \textit{dynamical similarity}. To measure this, we borrow a notion from theoretical computer science \citep{savage1998models, clarke2018model}, which defines systems as computationally equivalent if they are able to perfectly simulate one another. This means that a one-to-one mapping exists from the states of one system to the states of the other. If a pair of computationally equivalent systems are initialized to corresponding states and given an identical sequence of inputs, the sequences of states that they trace out will correspond perfectly. This notion was defined for discrete systems, but has been generalized to continuous systems like ours as ``approximate bisimulation'' \citep{girard2011bisimulation, mikulik2020meta}. We quantify this by running the ground-truth system on a battery of held-out experiments, generating a dataset of their internal states, and training neural networks to map the states of the ground truth system to the states of the discovered models, and the other way around. We evaluate the quality of these mappings on a separate dataset of internal states, and report the normalized mean-squared errors averaged over the two simulation directions (see Appendix \ref{app:bisimulation} for details). On Q-learning, ATLAS's bisimulation score after 100 steps matched that of random after 500 steps. On both Q-Learning and Leaky Actor-Critic, the bisimulation error was significantly lower for models trained on ATLAS-designed experiments, compared with random, after 100 experiments \ref{fig:eval}, right). These results hold for simulation error in each direction separately (Figure \ref{fig:ATLAS_vs_random_bisimulation} in Appendix \ref{app:bisimulation}). In line with the results on structure recovery, ATLAS was competitive with the expert-designed baseline on Q-learning, and even surpassed it on Actor-Critic, allowing for significantly lower bisimulation error by 100 experiments (Figure \ref{fig:handcrafted_vs_ATLAS}, right).

\section{Discussion}
\label{sec:discussion}

We introduce ATLAS, a pipeline for discovering mechanistic models through interpretable experimental design. Uniquely, ATLAS drives its optimal experimental design and ensemble-based active learning using a class of data-driven interpretable models.
We compare ATLAS to random experimentation on the problem of recovering the ground truth structure of reinforcement learning agents, and find that ATLAS is able to successfully recover mechanistic models that behaviorally, structurally, and computationally match the ground truth  using an order of magnitude fewer experiments. We observe that ATLAS selects intriguingly structured experiments that bear only slight resemblance to the kind of experiments researchers employ. Finally, we compare to a class of expert-designed experiments and find that ATLAS closely matches or outperforms that baseline.

While ATLAS is a general framework, we focused on the problem setting of recovering reinforcement learning algorithms for a few reasons. 
First, understanding the response of humans and animals (and artificial agents) to positive feedback is a problem of scientific, clinical, and societal importance, and one that has been a longstanding problem in cognitive science \citep{niv2009reinforcement, sutton2018reinforcement}.
Second, and more pragmatically, there has been substantial work developing data-driven interpretable modeling to reward-guided learning behaviors \citep{miller2018predictive,ji2025discovering,castro2025discovering,eckstein2026hybrid}. As our method relies crucially on such models, it makes sense to test in a domain where these models have been well tested.
Third, we wanted to test our method on problems with a known ground truth before extending to unknown systems. 
Future work with ATLAS would include applying it to domains outside of reinforcement learning and to experimental data collection in real world settings. 

There are a number of limitations to the methodology in its present form. In our setup, the rate-limiting step is training ensemble models, as these models are trained from scratch on each step. Future work might consider finetuning approaches that update the existing ensembles more efficiently. 
Using ATLAS with different methods for interpretable modeling also merit future work \citep{ji2025discovering,castro2025discovering}.
ATLAS uses an evolutionary approach to selects experiments greedily with respect to the EIG objective. Future work might optimize a batch of experiments \citep{kirsch2019batchbald}, or leverage RL to maximize \textit{cumulative} information gain \citep{treloar2022}. Although optimizing \textit{sequences} of experiments is computationally expensive, amortization \citep{foster2021deep} makes non-myopic design practical.


Accelerating data collection is a vital challenge given the ubiquity of data-driven modeling in AI. In many lines of scientific investigation, within and beyond cognitive science, data collection is the rate-limiting step, owing to the time and cost of real-world interventions. Our work addresses this for the problem of experimentally distinguishing and optimizing interpretable mechanistic models as opposed to blackbox predictive models. Such models play a vital role in cognitive science and beyond for instantiating specific, high-level hypotheses about the underlying mechanisms driving observed scientific process.
Our work is an important step for developing methods that accelerate the design of experiments that optimally uncover information about the underlying process while maintaining structure and interpretability. 

\section*{Acknowledgments}
We thank Will Dabney, Andre Barreto, Diana Borsa, Bernardo Avila Pires, Kalesha Bullard, and Martin Engelcke for helpful discussions, and the rest of the Neuroscience Lab and Agency Team at Google DeepMind for the supportive research environment. We also thank Will Dabney and Hado van Hasselt for providing feedback on earlier versions of the manuscript.

\bibliographystyle{unsrtnat}
\bibliography{references}

\newpage
\appendix
\section{Appendix}
\label{sec:app}

\subsection{Robustness of Experiment Optimizer}
\label{app:experiment_optimizer_seed_robustness}

We analyzed the robustness of the Experiment Optimizer on the example problem of distinguishing two Q-learning agents with different learning rates: $Q_{\text{Medium}}$ with learning rate $\alpha=0.2$ and $Q_{\text{Very Slow}}$ with learning rate $\alpha=0.05$ (corresponding to the first pair of agents shown in \ref{fig:experiment_optimizer_timescales}). In this analysis, we evolved $1{,}000$ independent samples starting from different seeds. All samples converged within $2{,}000$ steps. Within 10 subsets of 100 independent samples (considered as 'run' in our analysis), the best EIG score converges to the same value within $<0.1\%$ coefficient of variation. The best experiments on the 10 runs show remarkable consistency: they are arm-invariant versions of the same qualitative design. Each of the 10 best experiments has a partially-overlapping block structure, in which first only one choice is rewarding, then both are, then only the other choice is rewarding, then both are again, and finally only the first. Each optimal experiment includes several anomalous trials which deviate from the block structure. This confirms that 100 samples are sufficient for robust convergence in this problem setting.

\begin{figure}[h!]
  \centering
  \includegraphics[width=1.\textwidth]{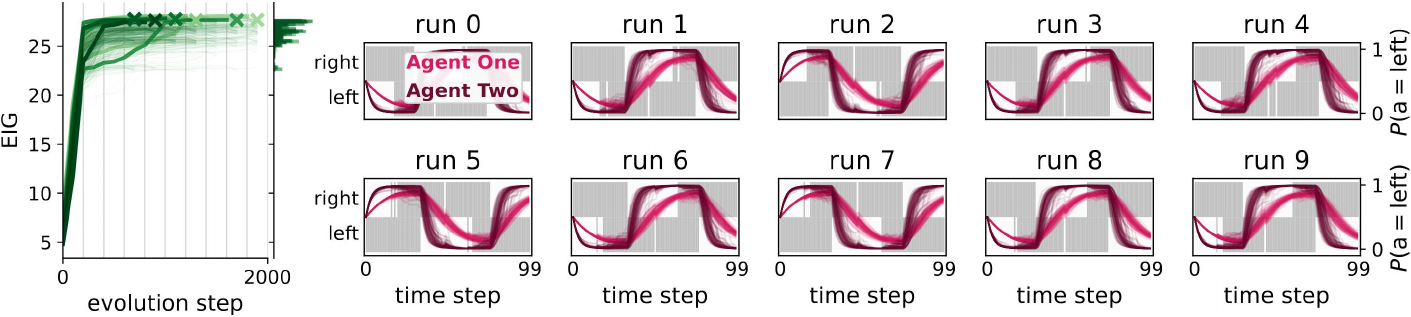}
\caption{\textbf{Robustness of optimized experiments for Q-Learners with different learning rates.} We optimized binary reward matrices to distinguish a $Q_{\text{Medium}}$ with learning rate $\alpha=0.2$ from $Q_{\text{Very Slow}}$ with $\alpha=0.05$. \textbf{(Left)} The expected information gain (EIG) converges in $<2{,}000$ evolution timesteps. Each line corresponds to one of 1,000 optimization runs, and different colors indicate subsets of 100 runs. Thick lines are the best experiment from each subset as evaluated on separate batches $B=2 \times 1024$, and crosses represent the score of this experiment on an additional, larger batch $B'=10 \times 1024$. Vertical lines represent convergence checks. \textbf{(Right)} Each panel displays the best optimized experiment from each subset of the runs. Experiments consist of a binary reward matrix where gray indicates that a reward of 1 will be given, and white indicates a reward of 0. Superimposed on each experiment is a batch of $N=100$ trajectories generated from each agent, with the action probabilities from $Q_{\text{Very Slow}}$ in pink, and from $Q_{\text{Medium}}$ in burgundy. The subsets identify highly similar experiments, with a partially overlapping reward block design.}

  \label{fig:experiment_optimizer_seed_robustness}
\end{figure}

\subsection{Agents}
\label{app:agents}

The data generating agents used in our experiments were two simple reinforcement learning agents a Q-learning agent and an actor-critic agent \citep{sutton2018reinforcement}.

\subsubsection{Q-Learning Agent}

The Q-learning agent maintains a latent variable $Q$ associated with each available action, and updates these following a reward according to:
\begin{equation}\label{eq:q_update_chosen}
    Q_{t+1}(a_t) = (1-\alpha)Q_t(a_t) + \alpha r_{t}
\end{equation}
and
\begin{equation}\label{eq:q_update_unchosen}
    Q_{t+1}(a \neq a_t) = Q_t(a \neq a_t)
\end{equation}
where $\alpha$ is a learning rate parameter set to $\alpha=0.1$. The Q-learning agent makes choices according to: 
\begin{equation}\label{eq:q_action_prob}
    a_t \sim \text{Softmax}(\beta Q_t)
\end{equation}
where $\beta$ is a softmax inverse temperature parameter controlling the degree of randomness in decision-making (set to $\beta=5$).

\begin{figure}[h!]
  \centering
  \includegraphics[width=.1\textwidth]{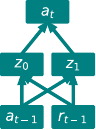}
  \caption{\textbf{The computational graph of Q-learning.} Viewed as a computational graph, the Q-learning agent with $A=2$ has two input nodes (previous choice and previous reward), two internal state nodes ($Q_1$ and $Q_2$) and one output node (next choice probability). This computational graph is moderately sparse: both inputs directly affect only two state variables, and those state variables directly affect the output but not each other.}
  \label{fig:qlearning_graph}
\end{figure}

\subsubsection{Leaky Actor-Critic Agent}

The leaky actor-critic agent maintains a set of action preferences $\theta$ for each available action, updated via a gradient bandit algorithm with a forgetting mechanism, as well as a running estimate of the recent reward rate, $v$.

The critic updates its value estimate $v$ following a reward $r_t$ according to:$$ v_{t+1} = (1 - \alpha_c) v_t + \alpha_c r_t $$where $\alpha_c$ is the critic's learning rate parameter, which we set to $\alpha_c=0.3$.This value serves as a baseline to compute the reward prediction error:$$ \delta_t = r_t - v_t $$The actor makes choices based on its current preferences $\theta_t$ by applying a softmax function to determine the probability of choosing action $a$:$$ \pi_t(a) = \frac{\exp(\theta_t(a))}{\sum_{a'} \exp(\theta_t(a'))} $$Following an action $a_t$, the preferences are updated using the prediction error $\delta_t$, an actor learning rate $\alpha_a$ (set to $\alpha_a=1.0$), and a forgetting rate parameter $\alpha_f$ (set to $\alpha_f=0.05$). The preference for the chosen action $a_t$ is updated according to:$$ \theta_{t+1}(a_t) = (1-\alpha_f)\theta_t(a_t) + \alpha_a \delta_t (1 - \pi_t(a_t)) $$and for the unchosen action $a \neq a_t$:$$ \theta_{t+1}(a \neq a_t) = (1-\alpha_f)\theta_t(a \neq a_t) - \alpha_a \delta_t \pi_t(a \neq a_t) $$

\begin{figure}[h!]
  \centering
  \includegraphics[width=0.1\textwidth]{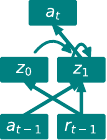}
    \caption{\textbf{The computational graph of Leaky Actor-Critic.} Viewed as a computational graph, the leaky actor-critic agent with $A=2$ has two input nodes (previous choice and previous reward), two internal state nodes (the critic, denoted by $z_0$, which depends on previous reward only, and the actor, denoted by $z_1$, which depends on the critic, the previous reward, and the previous action) and one output node affected only by the critic (next choice probability). 
    }
  \label{fig:actorcritic_graph}
\end{figure}

\subsection{DisRNN Hyperparameters and Fitting}
\label{app:disrnn_sweep}

\subsubsection{In ATLAS}

On every ATLAS cycle, we train DisRNNs on a range of hyperparameters to generate candidate models, which are then filtered to produce the ensemble. 
Specifically, we fit a total of 128 networks sweeping the DisRNN penalty parameter $\beta$ between between $10^{-4}$ and $10^{-1}$ and fitting four networks with different random seeds for each level of the penalty. We train the networks for 10,000 steps.

We evaluate each set of hyperparameters by computing a two-fold cross-validated likelihood score $L^{\text{cv}}$. We then obtain a single model for each set of hyperparameters by fitting a new network to the entire dataset. 
This results in a large set of models with very different tradeoffs between quality-of-fit and simplicity (Figure \ref{fig:ensemble_selection}).

\subsubsection{In evaluation}

To find the best-fitting model given a dataset produced by ATLAS, we fit a total of 64 networks sweeping the DisRNN penalty parameter $\beta$ between between $10^{-5}$ and $10^{-1}$ and fitting two networks with different random seeds for each level of the penalty. We train the networks for 100,000 steps.

Similarly to our model selection strategy within ATLAS, the best-fitting model is obtained by first selecting the set of hyperparameters that score highest in $L^{\text{cv}}$, and then fitting a network to the entire dataset using those hyperparameters.

\subsection{Battery of Held-out Experiments}
\label{app:heldout_experiments}

The held-out test set consists of a comprehensive grid of two-armed drifting bandit tasks. To ensure a robust evaluation across diverse environmental volatilities and learning timescales, the battery is generated by sweeping across combinations of initial reward probabilities, Gaussian random walk drift rates, and random seeds. Specifically, we generate a grid consisting of 6 linearly spaced initial reward probabilities between 0 and 1 for each of the two arms, 6 drift rates uniformly spaced between 0 and 0.3, and 6 random seed instantiations per combination. This yields a diverse suite of 1,296 unique experiments.

\subsection{Bisimulation Analyses}
\label{app:bisimulation}

We compute a bisimulation score for each of our discovered models by first learning mappings between their internal states and those of the ground-truth models. For this training phase, we simulate the ground-truth agent on the battery of heldout experiments (Appendix \ref{app:heldout_experiments}) to extract a trajectory of latents states, as well as to generate a dataset actions and rewards. We then pass the dataset of actions and rewards to the discovered model to extract its corresponding latent states.

\begin{figure}[!h]
  \centering
  \includegraphics[width=.6\textwidth]{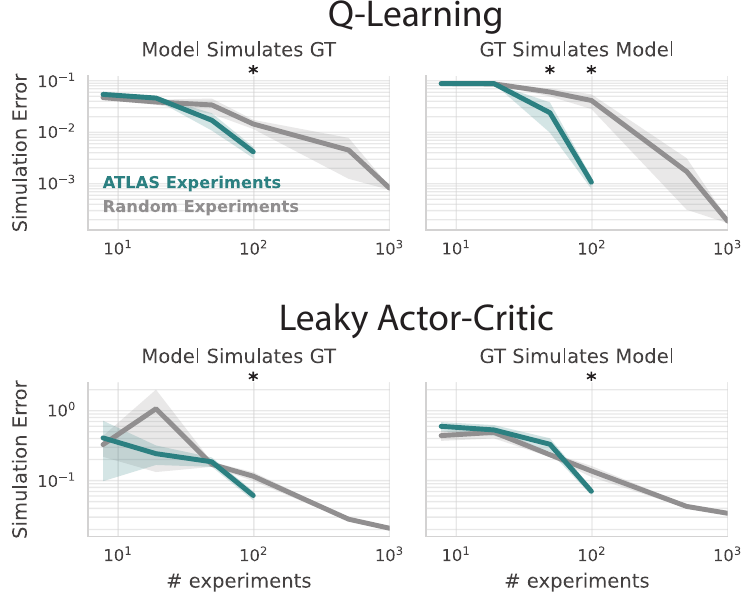}
  \caption{\textbf{Sample efficiency on bisimulation for ATLAS-designed versus random experiments}. We compare 8 independent runs of ATLAS to 10 runs of random experimentation. \textbf{(Left)} State prediction MSE for the ground truth agent (GT; Q-learning or Leaky Actor-Critic) simulated in the model. \textbf{(Right)} State prediction MSE for the discovered model simulated in the ground truth. Asterisks indicate significance ($p < 0.05$, Welch's one-tailed t-test).}
  \label{fig:ATLAS_vs_random_bisimulation}
\end{figure}

Next, we train two small feedforward neural networks: one to map the discovered model's state (e.g., recurrent unit activations) onto the ground-truth agent's state (e.g., Q-values), and the other to map the ground-truth agent's state onto the discovered model's state. Both mapping networks are multilayer perceptrons (MLPs) comprising 8 hidden layers with 32 hidden units each. These networks are trained with a batch size of 128 and a learning rate of $1 \times 10^{-4}$ for 5,000 steps.

Once these mappings are trained, we can "simulate" one system within the other: mapping an initial state across systems, applying the target's transition dynamics, and mapping the predicted states back. To evaluate fidelity, we generate 10 independent 32-timestep evaluation sequences and record the state traces for both systems.To ensure our evaluation is symmetric and invariant to the scale of different state spaces, we compute the final bisimulation score bidirectionally. First, we calculate the raw mean squared error (MSE) between the true and predicted traces in both directions (the discoevred model simulated in the ground-truth model and vice versa), averaged across time and state dimensions. We then normalize each raw MSE by dividing it by the mean standard deviation of its target system's states (computed across the time and batch dimensions, and bounded below by $10^{-8}$ to prevent division by zero). The final bisimulation score is the arithmetic mean of these two normalized MSEs (Figure \ref{fig:ATLAS_vs_random_bisimulation}), yielding a balanced, scale-invariant measure of algorithmic equivalence between the systems.

\subsection{Softmax Ensemble Sampling}
\label{app:ensemble_selection}

Default ATLAS explores hypotheses by sampling from an \textit{offset softmax} distribution over the cross-validated scores.

Let $\mathcal{M}$ denote the candidate models generated in Section \ref{subsec:hypothesis_generation}. For each model $m_i \in \mathcal{M}$ with a normalized cross-validation likelihood $L_{i}^{\text{cv}}$, we compute the weight:
\begin{equation}
    w_i = 1 + \exp\left(\frac{L_{i}^{\text{cv}} - c}{\tau}\right)
\end{equation}
where the baseline $c=0.5$ is set to match at-chance performance, and the temperature $\tau$ is a hyperparameter that controls the sharpness of the distribution. The $+1$ floor acts as a uniform mixture prior, guaranteeing a non-zero selection probability for the lowest-performing models. We then use the distribution $p_i = w_i / \sum_{j \in \mathcal{M}} w_j$ to sample $K$ models from $\mathcal{M}$ without replacement, forming our final ensemble $\mathcal{E}$. In our experiments we set a low temperature of $\tau=0.01$.

\begin{figure}[h!]
  \centering
  \includegraphics[width=1.\textwidth]{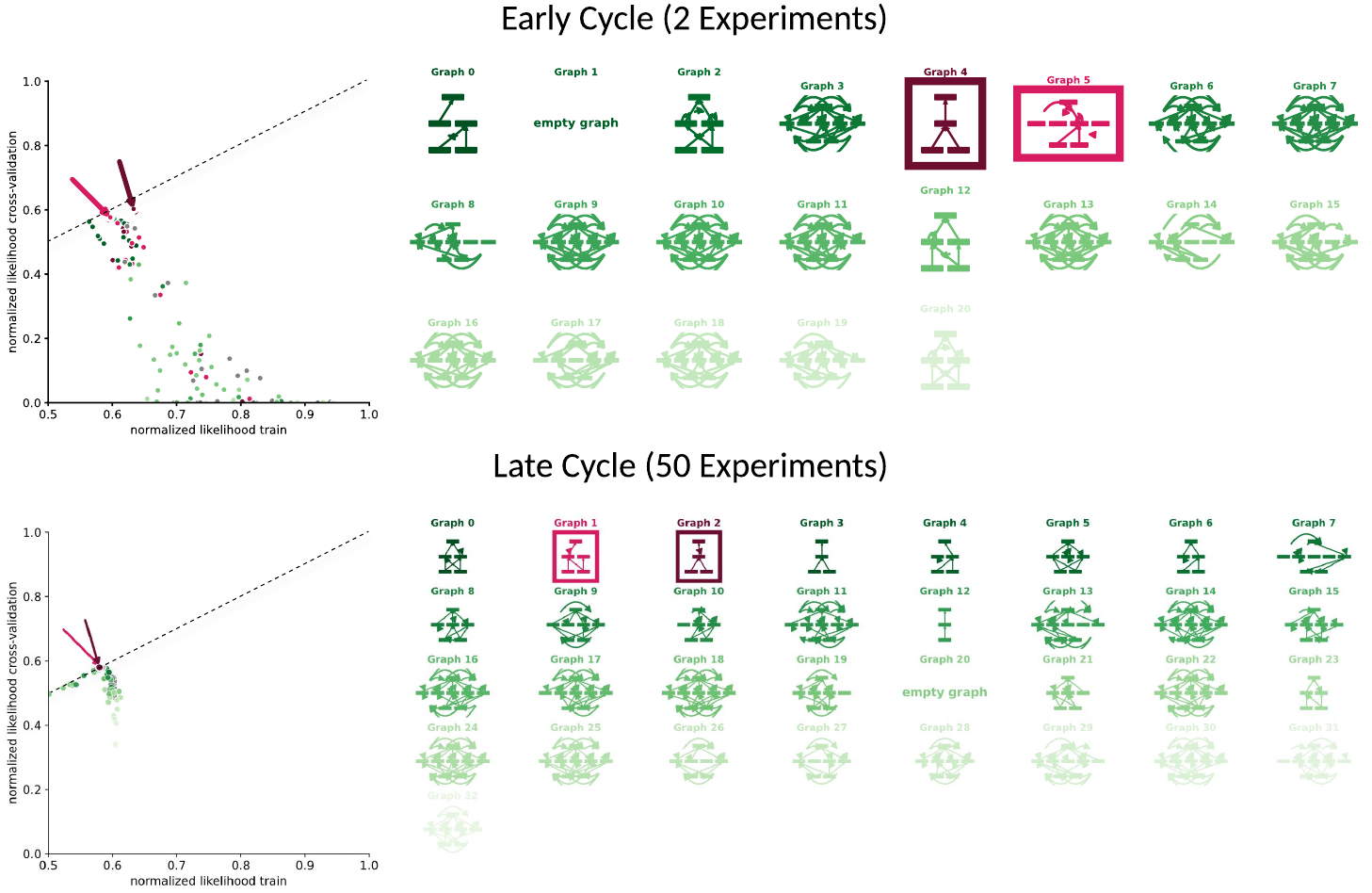}
  \caption{\textbf{Examples of early exploration using softmax ensemble selection.} \textbf{(Left)} The performance of the entire sweep of networks is plotted. The dashed line represents the unity line. Shades of green mark membership to one of the computational graph clusters that make up 80\% of the networks. (Other networks belonging to graph clusters with low counts are marked in gray). Darker shades mark higher average cross-validated normalized likelihood within the cluster. The two chosen ensemble members are marked in pink and burgundy and indicated with arrows. \textbf{(Right)} The set of unique graphs that cluster 80\% of the networks is shown in decreasing order of their cross-validated normalized likelihood, and the two chosen ensemble members are marked in pink and burgundy. In effect, softmax ensemble selection explores more on early cycles when models are closer to chance performance (Top), and more greedily later when more models have better performance (Bottom), as is evident in the selection of high-performing models nearer the start of the list later in training.}
  \label{fig:ensemble_selection}
\end{figure}

The rationale behind the offset softmax sampling is to focus experiment design on distinguishing the best models unless there are too few experiments to support any good models. The softmax rule implements this principle, effectively exploring more among the many overfit models on early cycles and preferring the top-scoring models on late cycles (Figure \ref{fig:ensemble_selection}).

\subsubsection{Robustness to Ensemble Selection Strategies}
\label{app:robustness_ensemble_selection}
The Active Learning literature describes a variety of ensemble selection heuristics, yet identifying the optimal approach for a given application remains an open challenge. We evaluated ATLAS's sensitivity to the ensemble selection strategy against other approaches adapted from the this literature on Q-learning agent recovery. Alongside ATLAS's default sampling softmax strategy, we evaluated a uniform sampling strategy, which effectively yields a discrete approximation of the Bayesian posterior necessary for information-theoretic acquisition functions like BALD \citep{beluch2018power, houlsby2011bayesian}. Furthermore, we evaluated a top-$k$ selection strategy, which acts as a practical proxy for sampling from the version space in Query-by-Committee approaches \citep{seung1992query}.

Overall, there was no significant difference among the three different strategies when testing the hypothesis that at least one of the strategies differed from the others, on any of the three metrics (all $p > 0.05$, One-Way ANOVA and Chi-Square tests). However, BALD-ATLAS converged on the ground truth earlier and by 50 experiments, it had a significant advantage over the default softmax-ATLAS, as revealed by Welch's one-tailed t-test ($p < 0.05$). This is a positive indication of ATLAS's robustness to different ensemble selection strategies.

\begin{figure}[h!]
  \centering
  \includegraphics[width=1.\textwidth]{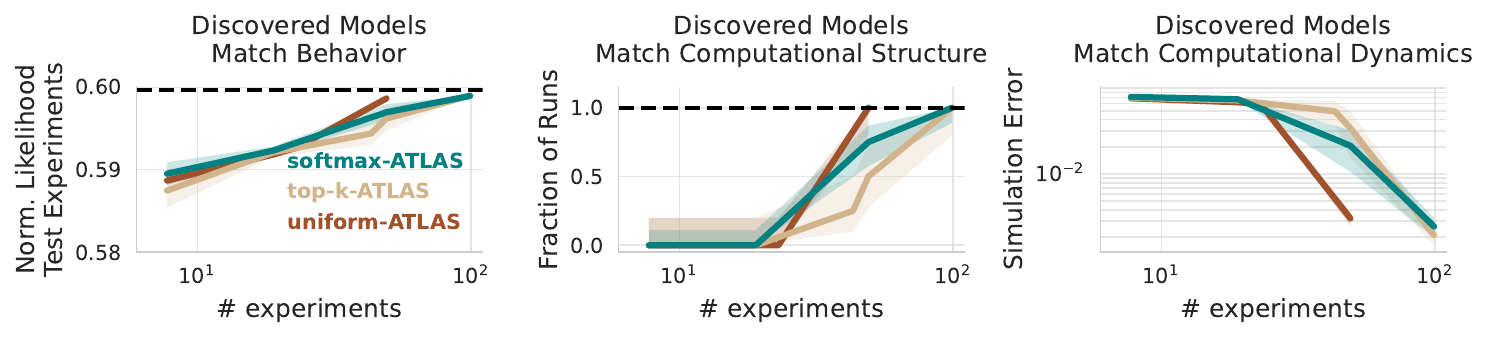}
  \caption{\textbf{ATLAS performance is robust to ensemble selection strategies.} On the task of recovering the Q-learning agent, the differences among the three ensemble selection strategies were not significant on any metric (all $p > 0.05$, One-Way ANOVA and Chi-Square tests).}
  \label{fig:qlearning_softmax-ATLAS_vs_BALD-ATLAS_vs_top-k-ATLAS}
\end{figure}

\subsubsection{ATLAS with GRU ensembles}
In order to better understand the effect of DisRNNs models on experimental design, we replaced the DisRNN models in the ensemble with vanilla gated recurrent units (GRUs), class of models shown to be effective at learning behavior models \cite{dezfouli2020adversarial}. We induced diversity in the pool of candidate models, and therefore in the ensembles, by sweeping the number of hidden layer units from 1 to 32, analogously to the effect of sweeping the penalty of DisRNNs.

\begin{figure}[h!]
  \centering
  \includegraphics[width=1.\textwidth]{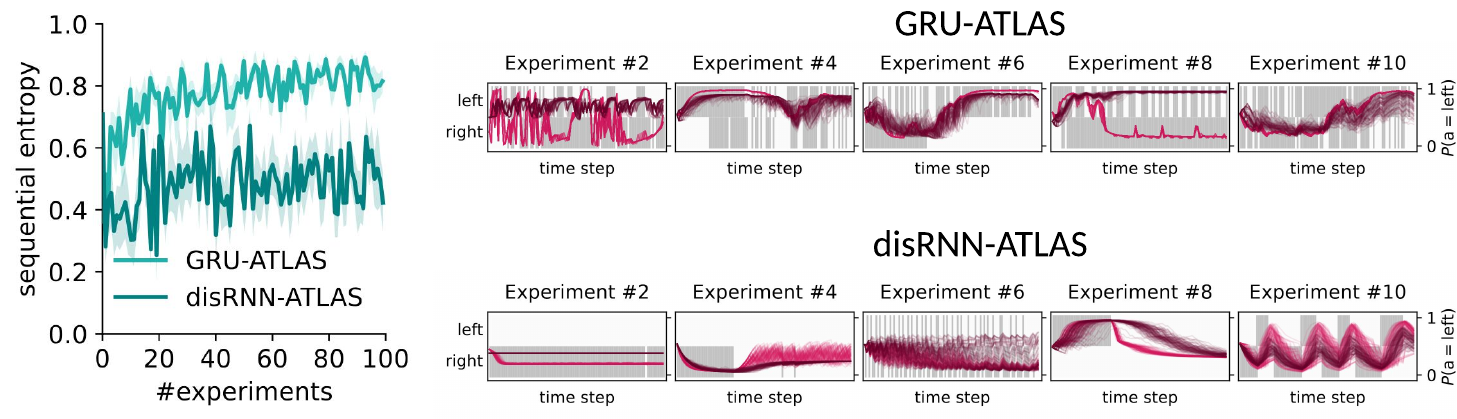}
  \caption{\textbf{DisRNN-ATLAS designs more structured experiments than GRU-ATLAS.} \textbf{(Left)} There is a marked and stable difference in the sequential entropy of experiments designed by GRU-ATLAS and DisRNN-ATLAS across 8 seeds. \textbf{(Right)} Example experiments and simulated trajectories from the two ensemble members are shown from the first 10 experiments on an arbitrarily chosen seed (seed 2) for GRU-ATLAS (Top) and DisRNN-ATLAS (Bottom).}
  \label{fig:experiment_entropy_GRU-ATLAS_vs_disRNN-ATLAS}
\end{figure}

Experiments designed by GRU-ATLAS show far less temporal structure (quantified by sequence entropy, see Appendix \ref{app:experiment_entropy}) than those designed by DisRNN-ATLAS (Figure \ref{fig:experiment_entropy_GRU-ATLAS_vs_disRNN-ATLAS}).

\begin{figure}[h!]
  \centering
  \includegraphics[width=1.\textwidth]{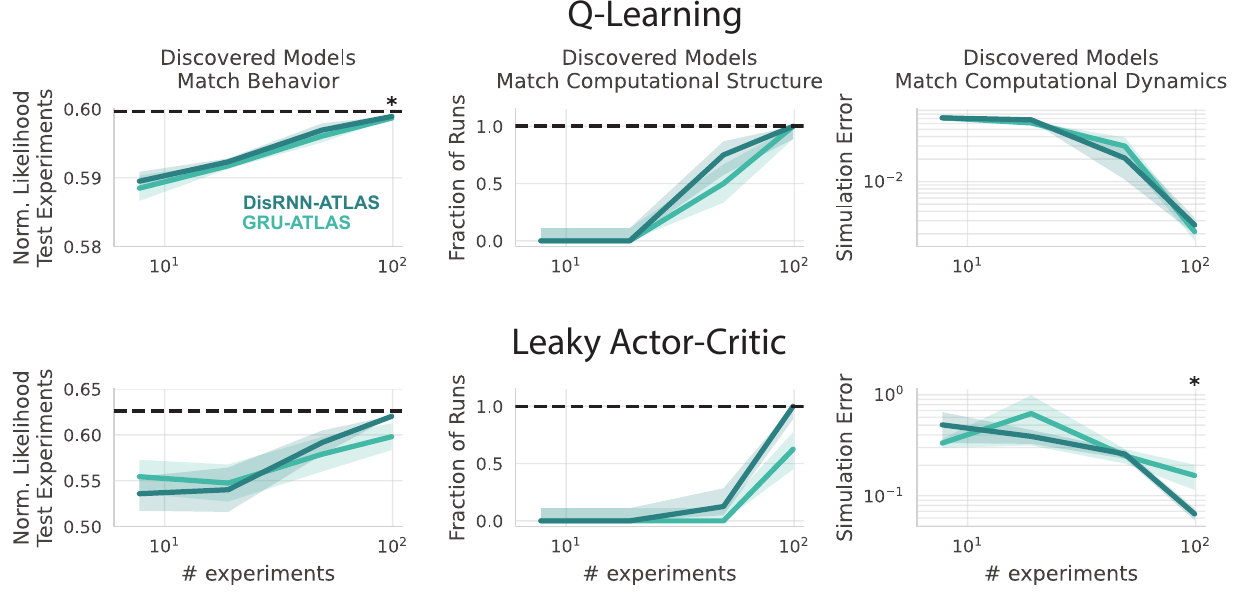}
  \caption{\textbf{GRU-ATLAS is on par with DisRNN-ATLAS}. We compare 8 independent runs of GRU-ATLAS and DisRNN-ATLAS on recovering Q-learning (Top) and leaky actor-critic (Bottom). \textbf{(Left)} Performance of the best-fitting models at predicting agent behavior in held-out experiments. Solid lines indicate the mean performance across independent runs, while the shaded regions represent $\pm 1$ standard error of the mean (SEM). Asterisks indicate significant advantage of DisRNN-ATLAS over GRU-ATLAS ($p < 0.05$, of Welch's one-tailed t-test). The horizontal dashed line represents the performance ceiling. \textbf{(Middle)} The probability of recovering the true underlying computational graph. Solid lines denote the fraction of successful recoveries on 8 seeds. Shaded regions represent the $68\%$  Wilson score interval. No significant differences were found (all $p > 0.05$, one-tailed Barnard's exact test). \textbf{(Right)} State prediction MSE for the ground truth agent simulated in the model. No significant differences were found (all $p > 0.05$, Welch's one-tailed t-test).}
  \label{fig:qlearning_GRU-ATLAS_vs_disRNN-ATLAS}
\end{figure}

GRU-ATLAS was also assessed in terms of its sample efficiency for recovering mechanistic models by fitting DisRNNs -- that is, while GRUs formed the ensembles for experiment design, DisRNNs were trained and evaluated on the data from those experiments. In terms of performance, GRU-ATLAS underperformed DisRNN-ATLAS albeit not significantly, except for a small performance decrement in terms of generalization (Figure \ref{fig:qlearning_GRU-ATLAS_vs_disRNN-ATLAS}).


\subsubsection{Experiment Entropy}
\label{app:experiment_entropy}

We quantified the first-order temporal structure in our experiment designs. Given a $T \times A$ design matrix $\mathbf{D}$, we calculated the conditional entropy of a reward given the immediately preceding reward for each action, averaged across all $A$ actions. The average sequential entropy is defined as:$$H_{seq} = \frac{1}{A} \sum_{a=1}^{A} H(\mathbf{D}_{t+1}(a) \mid \mathbf{D}_t(a))$$

where the conditional entropy for a given action's reward sequence is calculated using the empirical joint and marginal probabilities of successive reward pairs:$$H(\mathbf{D}_{t+1}(a) \mid \mathbf{D}_t(a)) = - \sum_{\mathbf{D}_t \in \{0,1\}} \sum_{\mathbf{D}_{t+1} \in \{0,1\}} P(\mathbf{D}_t, \mathbf{D}_{t+1}) \log_2 \left( \frac{P(\mathbf{D}_t, \mathbf{D}_{t+1})}{P(\mathbf{D}_t)} \right)$$

$P(\mathbf{D}_t, \mathbf{D}_{t+1})$ and $P(\mathbf{D}_t)$ refer to the empirical probabilities computed over the $T$ timesteps for action $a$.

\subsection{Compute resources}
\label{app:compute_resources}

On each cycle, an ATLAS run involves fitting 128 neural networks (32 penalty values and 4 seeds) and evolving experiments on 128 trajectories started from 128 different seeds. Each set of 8 independent ATLAS runs presented in the results requires a total of 1028 workers. One ATLAS cycle of generating novel hypotheses and designing an experiment takes $\sim30$ to $\sim60$ minutes. In our experiments, we ran ATLAS closed-loop for 100 cycles in $\sim48$ to $\sim96$ hours. 



\end{document}